\documentclass{article}

\usepackage{arxiv}

\usepackage[utf8]{inputenc} 
\usepackage[T1]{fontenc}    
\usepackage{hyperref}       
\usepackage{url}            
\usepackage{booktabs}       
\usepackage{amsfonts}       
\usepackage{nicefrac}       
\usepackage{microtype}      
\usepackage{lipsum}		
\usepackage{graphicx}
\usepackage{natbib}
\usepackage{doi}
\usepackage{comment}
\usepackage{amssymb}
\usepackage{amsmath}
\usepackage{subcaption}
\usepackage{placeins}

\newcommand{\argmax}{\operatornamewithlimits{argmax}}

\DeclareMathOperator{\onehot}{onehot}
\newcommand{\diff}[2]{\dfrac{\partial#1}{\partial #2}}

\title{Conditional Information Gain Trellis}

\author{ {\hspace{1mm}Ufuk Can Bicici}\\
	Bogazici University, Istanbul Turkey\\
	\texttt{can.bicici@boun.edu.tr} \\
	\And
	{\hspace{1mm}Tuna Han Salih Meral} \\
	Virginia Tech, Blacksburg VA \\
	\texttt{tmeral@vt.edu} \\
 	\And
	{\hspace{1mm}Lale Akarun} \\
	Bogazici University, Istanbul Turkey \\
	\texttt{akarun@boun.edu.tr} \\
}

\renewcommand{\shorttitle}{Conditional Information Gain Trellis}

\begin{document}
\maketitle

\begin{abstract}
 Conditional computing processes an input using only part of the neural network's computational units. Learning to execute parts of a deep convolutional network by routing individual samples has several advantages: Reducing the computational burden is an obvious advantage. Furthermore, if similar classes are routed to the same path, that part of the network learns to discriminate between finer differences and better classification accuracies can be attained with fewer parameters. Recently, several papers have exploited this idea to take a particular child of a node in a tree-shaped network or to skip parts of a network. In this work, we follow a Trellis-based approach for generating specific execution paths in a deep convolutional neural network. We have designed routing mechanisms that use differentiable information gain-based cost functions to determine which subset of features in a convolutional layer will be executed. We call our method  Conditional Information Gain Trellis (CIGT). We show that our conditional execution mechanism achieves comparable or better model performance compared to unconditional baselines, using only a fraction of the computational resources. \textbf{Important Note:} This paper has been accepted by Pattern Recognition Letters. For the peer-reviewed version of the paper, please visit: \url{https://doi.org/10.1016/j.patrec.2024.06.018}
\end{abstract}


\section{Introduction}

 Deep convolutional neural networks (CNN) have achieved high performance in many computer vision tasks, especially in image classification. One problem with these models is that they are heavily overparametrized, rendering them infeasible for edge computation and deployment on platforms like mobile phones and embedded devices.  Conditional computing, which disables a subset of the complete model based on a given sample, is one way to address this problem \cite{bengio2013deep}. This conditional execution can be used for forward calculations and/or back-propagation steps, effectively routing the given sample. For routing, different algorithms have been used in the literature. In this paper, we focus on the "Conditional Information Gain" algorithm 
 \cite{bicici2018conditional, BICICI2021108151}.
 The Conditional Information Gain Network (CIGN) builds a tree-structured network and introduces router elements into each non-leaf node. In CIGN, the router mechanisms trained with local information gain objectives try to maximize the information gain by learning an optimal split, such that the data routed into the subtrees become purer, in the sense that semantically similar classes are grouped together.  Convolutional filters in the corresponding sub-trees learn specifically discriminative representations for their respective data groups. This allows a reduction in the number of parameters in the tree branches. One problem with the tree structure is that misrouted samples can end up in sub-trees which are not specialized to discriminate them. We propose a new conditional computation model that remedies this misrouting problem.  

\begin{figure*}
\begin{center}
   \includegraphics[width=0.7\linewidth]{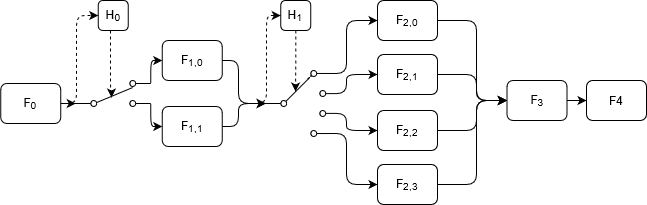}
\end{center}
   \caption{An example of the proposed CIGT architecture. The output of the first layer $F_{0}$ is used as input to router $H_{0}$. The output of $F_{0}$ follows the routing unit $k$ that $H_{0}$ decides. Then, the output of layer $F_{1,k}$ is used as input to router $H_{1}$. The output of $F_{1,k}$ follows the routing unit $j$ $H_{1}$ decides. And the output of $F_{2,j}$ follows the layers $F_{3}$ and $F_{4}$.}
   \label{cigt_lenet}
\end{figure*}

 This paper introduces a novel trellis-shaped model called Conditional Information Gain Trellis (CIGT),  that enables the recovery of misrouted samples. In a tree, there is only a single path between two nodes. In a Trellis structure, however, there are multiple paths between two nodes. If a sample is misrouted to a wrong computational unit, it will still have the chance to select the correct path in the next layer. CIGT uses the information gain-based routing mechanism in \cite{bicici2018conditional} in a DAG or Trellis structure (Fig. \ref{cigt_lenet}). 
We aim to create expert paths in the Trellis structure for different semantic groups in the given dataset. We divide the computational units into $N$ parallel groups (called $F$ units), which specialize in learning required representations for the data and a single $H$ unit, called a router, which learns a probability distribution over the computation units $F$. The local probability distributions on each block's $F$ units entail a global probability distribution on each root to leaf path in the Trellis structure and hence the model can be treated as a hierarchical mixture of experts model \cite{jordan1994hierarchical}. 
However, instead of using an Expectation Maximization based training approach, we  define routing mechanisms based on differentiable information gain objectives \cite{montillo2013entanglement, bicici2018conditional, BICICI2021108151}, which allow sparse execution of the computational blocks and group samples belonging to semantically similar classes together. When similar samples with small semantic differences follow the same root-to-leaf paths in the DAG structure, it can be hypothesized that the deep network making up the root-to-leaf path will learn to differentiate between the finer details of such samples, automatically converting the original multi-class problem to a simpler one. The probability distributions over each block's $F$ units are  trained with the previous block's corresponding local information gain losses. 

The paper is structured as follows: In section 2, we present a review of the  literature on conditional execution in deep network architectures. In Section 3, we  give the formulation of our CIGT model. Section 4 presents the experiments, in which we demonstrate our method's effectiveness on image datasets; MNIST,  Fashion MNIST and CIFAR 10. Section 5 concludes the paper.

\section{Related Work}
\subsection{Conditional Computing}

Conditional neural computation aims to reduce the computational cost during evaluation and training by selectively activating a subset of the network. It can try to create these subsets  by clustering with respect to a specific metric \cite{bengio2013deep}. The first work on conditional computation is 
in \cite{BengioStochastic}. It  proposes stochastic neurons, which can be switched on or off according to their outputs.  Blockout \cite{Blockout} proposes a generalization of Dropout: Blocks of the weight matrices in a neural network are set to zero stochastically. 
\cite{Bengio2015Conditional} calculates binary masks and applies them to the weight matrices. 
\cite{wu2017blockdrop} developed the ``Blockdrop" algorithm, which uses reinforcement learning to selectively drop some of the convolutional layers in a ResNet architecture. A similar work which drops layers using a gating mechanism is proposed in \cite{veit2018convolutional}. \cite{mcgill2017deciding} uses a multi-column and multi-layer CNN and uses different training procedures for routing samples. Skipnet uses probabilistic gating functions to skip layers completely \cite{wang2018skipnet}using a hybrid supervised-reinforcement learning setting in order to train the backbone network and the gates. \cite{herrmann2020channel} uses Gumbel-Softmax based channel selection in CNNs for conditional computation.

The methods discussed above are based on the principle of implicitly generating different network structures per sample. An alternative is to select an explicit hierarchical structure in the network a priori. 
\cite{ioannou2016decision} uses small routing networks for assigning weights to different computation paths and combining the top-k branches by weighted averaging.  
\cite{Liu} uses Q-Learning for training both the backbone network and the routers. With the exception of this, the method is similar to \cite{ioannou2016decision}. Genetic algorithms have also been used to learn the connections between different neural modules: Fernando et al. build a DAG structure of neural networks and their optimal connectivity is learned by a genetic algorithm \cite{PathNet}. 
The ``Branchynet" model \cite{branchynet} uses early exit branches based on an entropy threshold for classification. \cite{han2022learning} builds upon the early exit idea and introduces a weight learning scheme for early and late exiting samples, in order to bridge the gap between training and test behavior of such models.

Hierachical Mixture of Experts (HMoE) models lead to a natural interpretation for conditional computing \cite{jordan1994hierarchical}. With a gating mechanism, they assign probabilistic weights to different experts and combine the experts' decisions. Shazeer et al. propose a sparsely gated mixture of experts model, where the top-k experts with the highest selection weights are selected \cite{Shazeer}. Another line of approach which aims to reduce the computation load is pruning methods \cite{neill2020overview}. There is a clear distinction between conditional computing approaches and pruning methods: The latter treat the computation reduction as a training time problem; usually the network is pruned during the training or in a post-processing step and it is invariant to different inputs. Different approaches to pruning include Variational Dropout \cite{10.5555/3305890.3305939}, using sparsity regularizers \cite{7780649} and learning binary masks for weights \cite{tang2022automatic}. 

\subsection{Neural Network - Decision Tree Hybrids}

While CIGT uses a DAG structure, it uses ideas from ordinary decision trees. Neural Trees induct oblique decision trees, which contain perceptrons or MLPs as their splitting function in the non-leaf nodes \cite{foresti1998exploiting,rota2014neural}.
 Similar to common ID3 and C4.5 decision trees \cite{quinlan1986induction}, these algorithms use recursively growing decision trees. In the classical algorithms, at each split node, a class impurity measure is minimized using exhaustive search or random sampling. In Neural Trees, this is exchanged with a MLP based splitting algorithm which is trained by numerical optimization towards an optimal split configuration.

The Deep Neural Decision Forests  \cite{kontschieder2015deep}  replaces the softmax classifier layer of a deep neural network with differentiable decision trees. Interpreting the weight connections of a neural network as decision weights and building decision trees by using this information is another approach in this category of methods \cite{biau2016neural}. 
 A work in this vein by Baek et al. \cite{Baek} concentrates on converting deep convolutional neural networks into decision jungles \cite{decisionjungle}, by defining layer-wise entropy losses on probability distributions defined on the response map activation strengths.
 \cite{WangJournal} defines a tree structured neural network. In the split nodes, the distance to the current hyper plane is concatenated to the original input and then sent into child nodes.

 

"Adaptive Neural Trees" uses probabilistic routing similar to \cite{kontschieder2015deep} in a tree-shaped network, but allows the neural tree to grow based on heuristics from a validation set \cite{pmlr-v97-tanno19a}. Conditional Information Gain Networks (CIGN) \cite{bicici2018conditional} utilize a tree structure and introduce router elements into each non-leaf node. The routers are trained with local information gain objectives, maximizing information gain to group semantically similar classes together. Later, in \cite{BICICI2021108151}, the CIGN model is enhanced by adding entropy-based thresholds to the decision nodes. This ensures that samples for which the routers are indecisive are sent into multiple expert paths, based on the router's probability distribution exceeding a given threshold. The entropy thresholds are determined using Bayesian Optimization in a post-processing step.

A recent survey on dynamic neural networks can be found at \cite{han2021dynamic} which treats conditional computation models as a sub-genre of dynamic networks.

\section{Method}
The CIGT algorithm builds upon a Trellis structure of computational units, an example of which is given in Fig. \ref{cigt_lenet}. A deep neural network is divided into blocks of convolutional or dense layers $F_i$.   $F_0$ contains low level abstractions shared by all layers. Block $l$ has $K_l$ parallel processing units, with $1/K_l$ times fewer parameters each. 
Although blocks can be heterogeneous, we opt to use identically sized units. Each unit can contain variable number of convolutional or fully connected layers, which have common deep neural neural network operations like Batch Normalization \cite{BatchNorm}, Dropout \cite{srivastava2014dropout}, and different types of non-linear activation. We call the $i$th computational unit at the $l$th block as $F_{l,i}$. When only a single $F_{l,i}$ unit is selected in each layer, a  sequence of computational nodes $(F_{L,i_L} \circ \dots \circ F_{2,i_2} \circ F_{1,i_1} \circ  F_{0})(x) $ makes for a single root-to-leaf expert defining a posterior distribution $p(y|Z,x,\theta)$ for the given data point $(x,y)$, where $\theta$ represents all neural network parameters. $Z$ is defined as the matrix indicating the expert selection. Row $l$  of $Z$ shows the expert selection setting for the corresponding block. This row, $Z_l$, is a binary vector, $Z_l \in \{0,1\}^{K_l}$. 

The routers are indicated as $H$ units and are trained with the information gain loss. They are  connected to the intermediate outputs of the $F$ units in the same layer and parametrize a probability distribution $p(Z_{l+1}|x,\theta,\phi)$ on the next layer of $F$ units. $\phi$ contains all parameters which transform the intermediate input into a probability distribution in the $H$ unit. 
These layer-wise distributions on the $F$ units constitute a probability distribution over all root-leaf expert paths in the model as:

\begin{equation}
\label{Z_Distribution}
p(Z|x,\theta,\phi) = \prod_{l=1}^{L} p(Z_l|x,\theta,\phi)
\end{equation}

For the sake of simplicity in the notation, we simply write $Z$ as the representation for all the $Z_l$ in every layer in the text, unless stated otherwise.

Overall, given a dataset $\{x_i,y_i\}_{i=1}^{N}$, the main classification objective of the mixture of experts model can be written as:

\begin{equation}
\label{MainLoss}
\hat{J}_{C}(\theta,\phi) = \dfrac{1}{N}\sum_{i=1}^{N}\log \sum_{Z} p(y_i|Z,x_i,\theta) \prod_{l=1}^{L} p(Z_{l}|x_i,\theta,\phi)
\end{equation}

Instead of the objective in Eq. (\ref{MainLoss}), we use the following lower bound of it using Jensen's Inequality:

\begin{align}
\label{LowerBound}
J_{C}(\theta,\phi) = &\dfrac{1}{N}\sum_{i=1}^{N} \sum_{Z} \log p(y_i|Z,x_i,\theta) \prod_{l=1}^{L} p(Z_{l}|x_i,\theta,\phi)\\
<= \hat{J}_{C}(\theta,\phi) := &\dfrac{1}{N}\sum_{i=1}^{n}\log \sum_{Z} p(y_i|Z,x_i,\theta) \prod_{l=1}^{L} p(Z_{l}|x_i,\theta,\phi) \nonumber
\end{align}

We also have the information gain losses which lead the data distribution to computational units in each layer $l$, $IG_l$, with which our global loss becomes:

\begin{equation}
\label{Global_cost}
L_{CIGT} = -J_{C}(\theta,\phi) - \lambda_{IG} \sum_{l=1}^{L}IG_l + \Omega(\theta,\phi)
\end{equation}

where $\lambda_{IG}$ determines the strength of the overall information gain losses and $\Omega$ is a regularizer on the network parameters.

\subsection{Local Information Gain Loss}
The $H$ units define probability distributions $p(Z_l|x,\theta, \phi)$ over the $F$ units of the next layer. The router parameters $\phi$ and network parameters $\theta$ are trained with the information gain losses defined using these local probability distributions. An important step towards making such information gain objectives differentiable has been taken by Montillo et al. \cite{montillo2013entanglement}, by changing the step function with a sigmoid. Later, Bicici et al. \cite{bicici2018conditional} proposed to incorporate a similar version of this information gain loss into a tree structured neural network with similar routing mechanisms. Similarly, we define the following joint probability distribution for each block $l > 0$:

\begin{equation}
\label{JointDistribution}
p(y,Z_l,x|\theta,\phi) = p(y,x)p(Z_l|x,\theta,\phi)
\end{equation}

Here, $p(y,x)$ is the empirical training data distribution. We define $p(Z_l|x,\theta,\phi)$ as a softmax over possible values of $Z_l$:

\begin{equation}
\label{Softmax}
p(Z_l=k|x,\theta,\phi) = \dfrac{\exp\{(w_k^T h_l(x) + b_k)/\tau\}}{\sum_{i=1}^{K_l}\exp\{(w_i^T h_l(x) + b_i)/\tau\}}
\end{equation}

The hyperplanes $\{w_k,b_k\}$ determine the decision boundaries at the routers and $h_l(x)$ represent the output of the transformations in the $H$ units. 
$\tau$ is a hyperparameter controlling the smoothness of the probability distribution. As $\tau \to \infty$, the probability distribution approaches to uniform and as $\tau \to 0$, it tends to $\onehot{\left( \argmax_{i}\left(w_i^T h_l(x) + b_i\right) \right)}$, which produces a one-hot vector, setting the  entry corresponding to the largest logit to one and the others to zero. The local information gain at layer $l$ 
is then defined as:

\begin{equation}
\label{InformationGain}
IG_{l} = \mathbb{H}\left[p(y)\right] -  \mathbb{E}_{p(Z_l|\theta,\phi)} \left[ \mathbb{H}\left[p(y|Z_l,\theta,\phi)\right]  \right]
\end{equation}

Here $\mathbb{H}\left[p(x)\right]$ is defined as the entropy of the probability distribution $p(x)$: $\mathbb{H}\left[p(x)\right]=-\sum_{x}p(x)\log p(x)$. The derivative of the information gain respect to a generic network parameter is calculated as in \cite{bicici2018conditional}.

Using the above information gain criterion as an optimization target may result in convergence to a local minimum, which 
yields an unbalanced sample distribution to the corresponding $F$ units. We want to avoid this case since it may cause some paths in the Trellis structure to "starve" in the sense that they receive few samples and do not receive sufficient error signals. We handle this problem by using the solution proposed in \cite{bicici2018conditional}, utilizing the information gain framework's mathematical properties. The information gain can be decomposed as in the following:

\begin{equation}
\label{IG_decomposed}
IG_{l} = \mathbb{H}\left[p(y)\right] + \mathbb{H}\left[p(Z_l|\theta,\phi)\right] - \mathbb{H}\left[p(y,Z_l|\theta,\phi)\right]
\end{equation}

The information gain is maximized when $\mathbb{H}\left[p(Z_l|\theta,\phi)\right]$ attains its largest and $\mathbb{H}\left[p(y,Z_l|\theta,\phi)\right]$ attains its smallest value. $\mathbb{H}\left[p(y)\right]$ is a function of the data distribution hence is not affected by the network parameters. We note that a higher $\mathbb{H}\left[p(Z_l|\theta,\phi)\right]$ means a more balanced distribution of the data to the $F$ units in block $l$. We can increase the significance of this entropy term by defining a hyperparameter $\lambda_{balance} > 1$ as done in \cite{bicici2018conditional} and the information gain loss becomes:

\begin{equation}
\label{IG_decomposed}
IG_{l}^{balanced} = \mathbb{H}\left[p(y)\right] + \lambda_{balance}\mathbb{H}\left[p(Z_l|\theta,\phi)\right] - \mathbb{H}\left[p(y,Z_l|\theta,\phi)\right]
\end{equation}

This modified cost will prefer minima with more balanced data distributions on the computation nodes. The information gain objective provides an elegant mechanism to remedy this load balancing problem (or "mode collapse" as named in \cite{NIPSModular}), compared to the additional proxy balancing losses in \cite{Shazeer, Bengio2015Conditional,Baek}. 


\subsubsection{Approximate Training}
The derivatives of the lower bound in Eq. (\ref{LowerBound}) with respect to $\theta$ and $\phi$ are given as:

\begin{align}
\label{DerivativeTheta}
\diff{J_{C}}{\theta} = &\dfrac{1}{N}\sum_{i=1}^{N} \sum_{Z} \diff{\log p(y_i|Z,x_i,\theta)}{\theta} \prod_{l=1}^{L} p(Z_{l}|x_i,\theta,\phi)\nonumber\\
&+\sum_{i=1}^{N} \sum_{Z} \log p(y_i|Z,x_i,\theta) \diff{\prod_{l=1}^{L} p(Z_{l}|x_i,\theta,\phi)}{\theta}
\end{align}

\begin{align}
\label{DerivativePhi}
\diff{J_{C}}{\phi} =
&\dfrac{1}{N}\sum_{i=1}^{N} \sum_{Z} \log p(y_i|Z,x_i,\theta) \diff{\prod_{l=1}^{L} p(Z_{l}|x_i,\theta,\phi)}{\phi}
\end{align}

While the first term of $\diff{J_{C}}{\theta}$ allows a simple Monte Carlo approximation by drawing samples $x_i \sim p(x)$ and $Z_i \sim p(Z|x_i,\theta,\phi)$, its second term as well as the derivative $\diff{J_{C}}{\phi}$ is not an expectation with respect to some distribution, hence computing them with Monte Carlo simulations is not possible. 

In general, the gradients of the type $\dfrac{\partial}{\partial w}\mathbb{E}_{p(x|w)}[f(x)]$ are problematic to compute. There are two main types of remedy in such cases. The first one is the REINFORCE \cite{Williams92REINFORCE} type score function estimators. Another approach is to replace the discrete sampling procedure with approximate distributions like "Gumbel-Softmax" (\cite{jang2016categorical}, \cite{maddison2016concrete}).

In the CIGT framework, we follow a different approach for training the MoE loss which involves stochastic nodes. We ignore $\diff{J_{C}}{\phi}$ in Eq. (\ref{DerivativePhi}) and the second part of  $\diff{J_{C}}{\theta}$ in Eq.  (\ref{DerivativeTheta}) completely. We call the remaining term the reduced gradient $\hat{\diff{J_{C}}{\theta}}$). The $H$ unit  parameters will be solely updated with respect to the information gain losses at each block, namely $ -\lambda_{IG} \sum_{l=1}^{L}\dfrac{\partial}{\partial \phi}IG_l$. The parameters of the computation units ($F$) will also be updated via information gain objectives. At each block $l$, the router units use the outputs of the computational units to build and sample from the probability distributions $p(Z_{1}|\hat{x}_{l},\theta,\phi)$ where $\hat{x}_{l}$ are the outputs for each corresponding $F$ unit. According to the samples $Z_{l}$, the corresponding $F$ units in the next block are invoked. This is equivalent to sampling a root-to-leaf path for every sample, evaluating and then backpropagating the corresponding error signal through the followed path. Since we only use the gradients of the main loss partially and depend on the information gain losses to train the router nodes, we call this training algorithm as  "Approximate Training". 
During  evaluation, the most likely $F$ unit is selected from the corresponding $H$ unit's output for a sample $x$. This expert path is the most appropriate one in the sense that it is trained with samples semantically similar to $x$, hence has a high discriminative power. Only a single path is evaluated during inference,  making the algorithm efficient compared to unconditional baselines. We refer the reader to the supplementary materials for visual examples of the semantic groups generated in the CIGT routing.

\section{Experiments}

We have conducted experiments on MNIST \cite{lecun1998gradient},
Fashion MNIST \cite{xiao2017/online} and CIFAR 10 \cite{cifar100} datasets. For MNIST and Fashion MNIST datasets, we have used LeNet based models as in \cite{bicici2018conditional}. For CIFAR 10, we have used a ResNet110 based model, which is comparable to the ones used in \cite{veit2018convolutional} and \cite{herrmann2020channel}.

Our experimentation methodology consists of creating baseline architectures and converting these architectures to CIGT architectures by adding information gain routing blocks and splitting convolutional layers into routes. In each forward pass, our CIGT network corresponds to a slimmer version of the baseline network and the routing blocks. For a fair comparison, we have used the same layer configurations as the baseline architectures in CIGT networks. 

\subsection{MNIST Experiments}
For  MNIST, we use a LeNet baseline, which has 20 feature maps with $5 \times 5$ convolutional kernels, followed by 50 feature maps of the same kernel size, followed by 500 and 10 dimensional dense layers. Our root-to-leaf experts are similar to ones in \cite{bicici2018conditional}, where we reduce the second convolutional layer's feature map count to 15 and reduce the following dense layer's dimensionality to 25. The CIGT structure has three routing blocks. The first block contains the first convolutional layer as in the baseline. The second block contains two units of convolutions with 15 feature maps. The third block contains four units of the dense layers. We call this structure as $CIGT-[1,2,4]$, corresponding to the number of routing units. The routers are just 16 dimensional dense layers. We have used SGD optimizer with a momentum of 0.9. The batch size is 125 and the total number of epochs in each training is 100. We pick $\lambda_{\mathit{balance}} = 2$. The other hyperparameters are used as in \cite{bicici2018conditional}, with the difference that we do not use a separate weight decay term for routers. Our tests compare the CIGT against the thick baseline, the single root-to-leaf expert (slim) and a CIGT with random routing instead of information gain. All our CIGT results are the average of 5 different runs. Our experiments show that our method achieves higher accuracy not simply by reducing overfitting with fewer parameters, as the slim version of the baseline cannot achieve the same results. Furthermore, our experiments with random routing demonstrate that our learned routing mechanism is more than just a dropout mechanism for convolutional filters. 
The results are shown in Table \ref{tab:mnistresult}. CIGT surpasses all baselines and the equivalent, but tree-based conditional model, CIGN \cite{bicici2018conditional}. We also compare CIGT with Adaptive Neural Trees \cite{pmlr-v97-tanno19a} and Deep Neural Decision Forests \cite{kontschieder2015deep}; which use the same LeNet model as their backbone on MNIST. CIGT attains better performance compared to these and uses fewer MAC operations per sample.

\begin{table}
\begin{center}
\begin{tabular}{|l|c|c|c|}
\hline
Method  & Avg Acc. & Avg \# MAC & Parameters\\
\hline\hline
LeNet (Thick) & 99.25\% & $4.018 \times 10^6$ & 1256080\\
LeNet (Slim)  & 99.20\% & $1.057 \times 10^6$ & 26695\\
CIGT-(R*) & 99.29\% & $1.057 \times 10^6$ & 92276\\
\textbf{CIGT-[1,2,4]} & \textbf{99.40\%} & $1.057 \times 10^6$ & 92276\\
CIGN \cite{bicici2018conditional, BICICI2021108151} & 99.36\% & $1.057 \times 10^6$ & 92276\\
ANT \cite{pmlr-v97-tanno19a} & 99.36\% &  $\approx5.0 \times 10^6$ & -\\
DNDF \cite{kontschieder2015deep} & 99.30\% & $\approx 4.0 \times 10^6$ & -\\
\hline
\end{tabular}
\end{center}
\caption{MNIST Results. Each sample in the CIGTs visits a network equivalent to LeNet (Slim) plus router blocks. CIGT-(R*) network uses an equivalent CIGT-[1,2,4] architecture but routings are made randomly.}
\label{tab:mnistresult}
\end{table}

\subsection{Fashion MNIST Experiments}


For Fashion MNIST tests, we have used a LeNet CNN model consisting of 3 convolutional and 3 fully connected layers as our baseline. The convolutional layers consist of 32, 64 and 128 convolutional filters respectively, with $5 \times 5$ filters and $2 \times 2$ maxpooling layers after each convolutional layer. Dense layer sizes are 1024, 512 and 10 respectively. We used the similar CIGT-[1,2,4] setting as in the MNIST experiments. We use the first convolutional layer of 32 feature maps as the first block. The second layer with 64 convolutional feature maps is split into two routes with 32 feature maps in each. We change the number of feature maps of the third convolutional layer to 32 and the dimensions of the following two fully connected layers to 128 and 64. 
The decision networks take as inputs the outputs of the first and second convolutional layers. Global Average Pooling is applied to these and then a 128 dimensional fully connected layer generates the input feature for the information gain routers.

In the experiments, we have used a SGD optimizer with a momentum of 0.9. We have trained each model in 125 epochs with a batch size of 125 and a starting learning rate of 0.01. In the first 25 epochs, the network is trained without conditional routing as a warm-up phase. This is similar to the training with annealing approach in \cite{BICICI2021108151}. The learning rate is halved at 27000. and 42000. iterations and multiplied by 0.1 at 52000. iteration. For regularization, we have used Dropout after each fully connected layer. We picked Dropout probability as $0.3$ and $\lambda_{balance} = 2$. We pick the routing loss coefficient $\lambda_{IG}$ empirically as $0.7$ in order to bring the classification and routing losses around the same magnitude. The Softmax temperature decay procedure is left as the same with the MNIST experiments.

\begin{table}
\begin{center}
\begin{tabular}{|l|c|c|c|}
\hline
Method  & Avg Acc. & Avg \# MAC & Parameters\\
\hline\hline
CNN (Thick) & 92.25\% & $6.522 \times 10^6$ & 1771018\\
CNN (Slim) & 91.65\% & $2.606 \times 10^6$ & 73418\\
CIGT-(R*) & 92.09\% & $2.659 \times 10^6$ & 294702\\
\textbf{CIGT-[1,2,4]} & \textbf{92.52}\% & $2.659 \times 10^6$ & 294702\\
CIGN \cite{BICICI2021108151}& 92.37\% & $3.154 \times 10^6$ & 368814\\
sMoE-C. $\lambda=0.9$ \cite{BICICI2021108151}& 92.52\% & $3.165 \times 10^6$ & 368814\\

\hline
\end{tabular}
\end{center}
\caption{Fashion MNIST Results. Each sample in the CIGTs visits a network equivalent to LeNet (Slim) plus router blocks. CIGT-(R*) network uses an equivalent CIGT-[1,2,4] architecture but routings are made randomly.}
\label{tab:fashionmnistresult}
\end{table}

All CIGT-related results in Table \ref{tab:mnistresult} are the average of 10 different runs with no augmentation. We can see that random routing in the CIGT structure provides a regularization effect, however, the actual information gain routing leads to a better result over random routing; demonstrating that the performance improvement is not solely caused by the reduction of overfitting. We also compare the CIGT model with the CIGN and sMoE-CIGN models of \cite{BICICI2021108151}, which use comparable architectures on Fashion MNIST. sMoE (Sparse Mixture of Experts) models enable multiple-path routing during inference, whenever the routers exhibit an entropy larger than a determined threshold, while our CIGT models are strictly using a single root-to-leaf path \cite{BICICI2021108151}. Despite that, the Trellis network attains an equivalent performance as the sMoE-CIGN model with the multi-path inference capability, at a lower inference cost.

\subsection{CIFAR 10 Experiments}
For CIFAR 10 experiments, we used a ResNet110 backbone, similar to \cite{veit2018convolutional, herrmann2020channel}. Our CIGT networks utilized [1,2,2] and [1,2,4] routing options. The first routing block (1 unit) contains 9 ResNet blocks, each with 16 feature maps and no bottleneck layer \cite{He_2016_CVPR}. The second routing block has 2 units, each with 27 ResNet blocks: the first 9 have 12 feature maps, and the remaining 18 have 16 feature maps after applying a $2 \times 2$ stride. The final routing block (2 or 4 units) consists of units with 18 ResNet blocks, with the first block using a $2 \times 2$ stride. We used Convolutional Block Attention Modules (CBAM) \cite{woo2018cbam} as routers, applying 3 layers of CBAM with a 4-reduction rate and $4 \times 4$ average pooling to calculate the information gain after passing through the units in a routing block.

\begin{table}
\begin{center}
\begin{tabular}{|l|c|c|c|}
\hline
Method  & Avg Acc. & Avg \# MAC & Parameters\\
\hline\hline
R-110 (Thick) & 93.39\% & $186.8 \times 10^6$ & 1730714\\
R-110 (Slim) & 91.76\% & $78.2 \times 10^6$ & 235218\\
CIGT-(R*) & 92.32\% & $78.2 \times 10^6$ & 644478\\
\textbf{CIGT-[1,2,4]} & \textbf{93.81}\% & $78.8 \times 10^6$ & 644478\\
\textbf{CIGT-[1,2,2]} & \textbf{93.26}\% & $78.8 \times 10^6$ & 475112\\
AIG-110 \cite{veit2018convolutional} & 94.24\% & $153.3 \times 10^6$ & $\approx 1.74 \times 10^6$\\
CSGS-110 \cite{herrmann2020channel} & 94.36\% & $121.5 \times 10^6$ & $\approx 1.74 \times 10^6$\\

\hline
\end{tabular}
\end{center}
\caption{CIFAR 10 Results. Each sample in the CIGTs visits a network equivalent to R-110 (Slim) plus router blocks. CIGT-(R*) network uses an equivalent CIGT-[1,2,4] architecture but routings are made randomly.}
\label{tab:cifar10result}
\end{table}


In our experiments with ResNet-110 based CIGT models, we use minibatches of size 1024. To further facilitate the generalization of routers, we split each minibatch into two: One-half of each minibatch goes through the usual CIFAR 10 augmentations of random crop and random horizontal flip; these are used for the training of the classifiers. The other half of the minibatch goes through RandAugment \cite{cubuk2020randaugment} augmentations, which select two random augmentations per image from a pool of augmentations. These more heavily augmented samples are used for training the routers with the information gain loss. We use an SGD optimizer, that starts with a learning rate of $0.1$. At $950.$ and $1350.$ epochs, we multiply the learning rate with $0.1$. The whole training lasts for $1750$ epochs. The first $350$ epochs are used as a warm-up; during these epochs, we keep training the routers, however, the routing decisions are done randomly. This allows each path in the Trellis structure to see samples from each class initially. We use $0.0005$ as the weight decay term. The balance coefficient $\lambda_{balance}$ for the CIGT-[1,2,2] model is left at $1$, but we set it $\lambda_{balance}=5$ for the CIGT-[1,2,4] model since we observe that routing into four paths in the last block might lead into load-balancing issues during the training. $\lambda_{IG}$ is set to $1$ during all experiments.

 We observe in Table \ref{tab:cifar10result} that  CIGT-[1,2,4] surpasses the performance of the Thick ResNet-110 baseline significantly, using only $41.8\%$ of the computation load and $37\%$ of the number of parameters. We observe a performance drop in CIGT-[1,2,2] variant compared to CIGT-[1,2,4]. In addition to the obvious extra routing capability in the latter model, setting $\lambda_{balance}=5$ in CIGT-[1,2,4] also seems to help the training to stabilize, leading to a further performance increase. Comparison with Adaptive Inference Graphs (AIG) \cite{veit2018convolutional} and Channel Selection with Gumbel Softmax (CSGS) \cite{herrmann2020channel}, presenting results with comparable ResNet-110 based models on CIFAR-10 is also provided. Both  report a higher performance compared to our best model. We attribute this gap to the different granularities of the competitor methods: 
 AIG can skip individual ResNet blocks and CSGS further chooses individual feature maps to ignore with the virtue of the Gumbel-Softmax sampling. In CIGT, we skip blocks of computation in their entirety. This allows for modular computation in a scenario where each block is evaluated on a distributed computation server. While this limits the search space for model sparsity, it helps us to make the inference more efficient: AIG-110 uses $82\%$ and CSGS-110 uses $65\%$ of the thick model's computation load, where we only need $41.8\%$ of it and can still surpass the thick model's performance significantly. Moreover, CIGT models tend to be more compact; both AIG-110 and CSGS-110 use the original thick model only to adjust it for conditional computation during inference time. 
 Our best model only requires $37\%$ of the parameters needed in the original ResNet-110. Additionally, these methods and CIGT are orthogonal to each other: Theoretically, it is still possible to apply the Gumbel-Softmax Channel Selection approach in \cite{herrmann2020channel} in a Trellis shaped CIGT model.

\section{Conclusion}
In this paper, we have proposed a Trellis-shaped CNN model, named Conditional Information Gain Trellis (CIGT) on which samples can be routed during both training and inference, using a routing network trained with information gain objectives. We have proposed methods and strategies for its training and inference.  We have conducted experiments on three popular datasets: MNIST, Fashion MNIST and CIFAR-10. We show that CIGT performs better compared to unconditional baselines with heavier computational burden and it also performs better or comparable to similar conditional computation methods in terms of accuracy and reduction of the MAC operations. Our future work will involve integrating inference time multiple-path routing strategies into the CIGT architecture for improving the method accuracy, as done for the tree-based CIGNs in \cite{BICICI2021108151}.

\section{Appendix}

This appendix contains the supplementary materials to the paper "Conditional Information Gain Trellis". The information gain based-routing in CIGT aims to group samples belonging to semantically similar classes together. When similar samples with small semantic differences follow the same root-to-leaf paths in the DAG structure,  the deep network making up the root-to-leaf path can learn to differentiate between the finer details of such samples, automatically converting the original multi-class problem to a simpler one. In this document, we show how the information gain based routing partitions data into semantically meaningful sub-parts in histograms computed on the test sets of Fashion MNIST, CIFAR-10 and MNIST datasets.

For each dataset, we show the routing structure used and provide histograms belonging to the most frequent classes routed to that block. The histograms are cut off  when the cumulative probability passes the $\%85$ threshold. In the routing unit images, each row shows five random samples from the corresponding class for that row, with their frequencies. The next group of figures show the exact number of test set samples routed to the blocks.

\subsection{Semantic Partitioning of Data}

\FloatBarrier

\subsubsection{Fashion MNIST}

\begin{figure}[!t]
\centering
\includegraphics[width=4.75in]{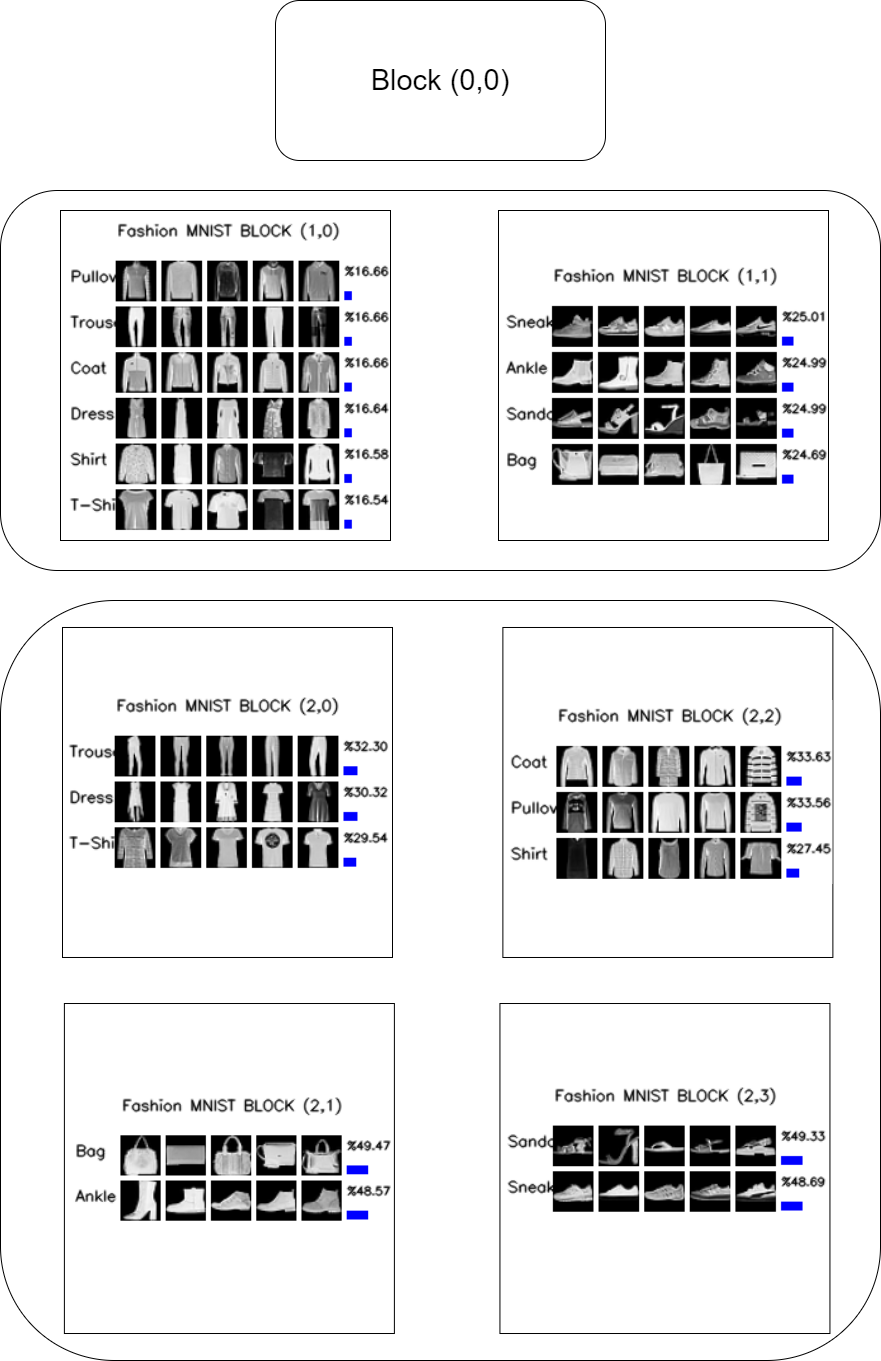}
\caption{The class distribution of Fashion MNIST samples into CIGT routing units.}
\label{FashionMNIST_CIGT}
\end{figure}

Fashion MNIST has a clear visible distinction between three groups: There are three footwear, six clothing classes, and there is  the Bag class, which does not strictly belong to these two groups. Figure \ref{FashionMNIST_CIGT} shows how these classes are grouped together in the [1,2,4] structure of the CIGT models we use. Histograms of computed on the test set have been used for demonstrating the routing distributions. It is clearly visible that in Block(1,0) and Block(1,1) the six clothing classes have been separated from the three footwear classes and the Bag class. In the next routing block, we have the Trouser, Dress and T-Shirt classes routed into Block(2,0). Both Trouser and Dress samples tend to resemble a rectangle with a large height and smaller width, as a broad visual abstraction. T-Shirt samples have usually short arms, that is visually similar to the upper body parts of the Dress category, hence it is logical that Dress and T-Shirt have been routed together. The Block(2,1) contains Bag and Ankle Shoe categories, dominantly. Both these categories resemble larger width and shorter height rectangles visually, so while the Bag class is not particularly similar to Ankle Boots in finer details, this lower level of visual similarity might lead to this particular grouping. In Fashion MNIST experiments a $\lambda_{balance}$ value of 2 is used, which encourages the routers to attain a balanced distribution of samples, hence this might have also contribute to the grouping of Bag with Ankle Boots, instead of grouping all footwear classes together while leaving the Bag class single. Block(2,2) has Coat, Pullover and Shirt classes most dominantly, which is a natural way of grouping upper body related clothing samples with generally longer arm parts. Finally Block(2,3) has the classes Sandals and Snekaers. Both are footwears and they tend to have shorter ankle parts that might be the reason of their clustering. As can be seen from these examples, the groups that have been generated by the CIGT structure have plausible semantic explanations.

\begin{figure}
     \centering
     \begin{subfigure}[b]{0.675\textwidth}
         \centering
         \includegraphics[width=\textwidth]{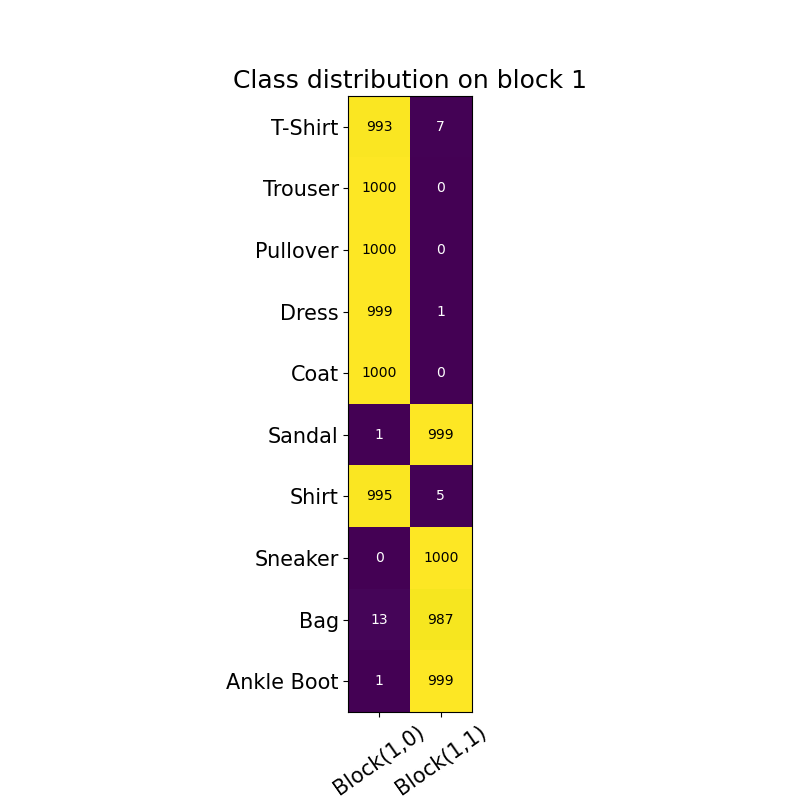}
     \end{subfigure}
     \hfill
     \begin{subfigure}[b]{0.675\textwidth}
         \centering
         \includegraphics[width=\textwidth]{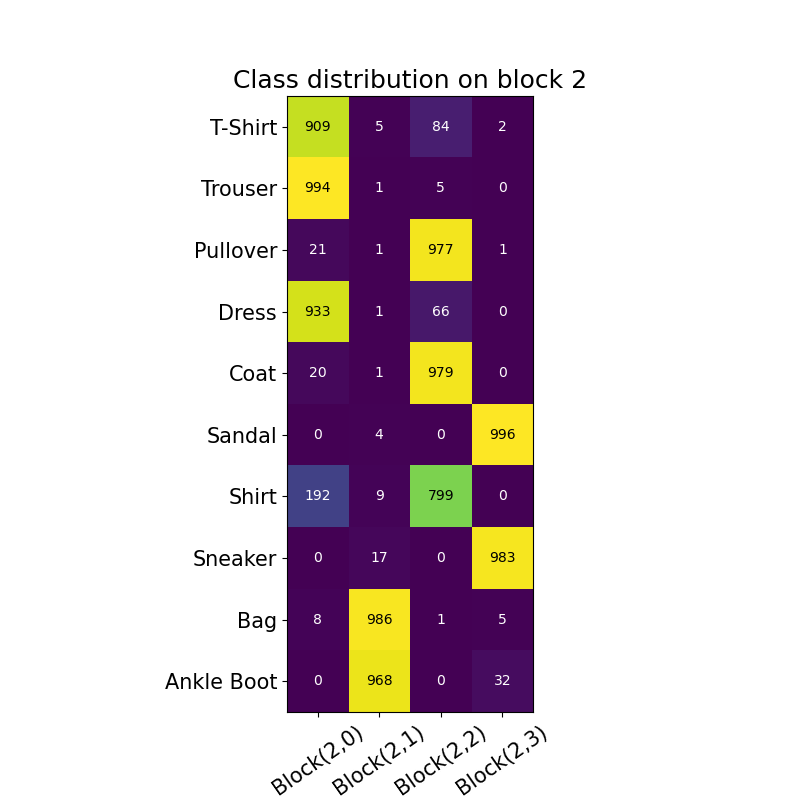}
     \end{subfigure}
        \caption{Fashion MNIST Test set sample distribution in each of routing blocks.}
        \label{fig:FashionMNIST_RoutingMatrices}
\end{figure}

Figure \ref{fig:FashionMNIST_RoutingMatrices} presents the numerical distribution of test samples from the same experiment across various computation units. An interesting observation is the slight tendency of the Shirt class to be routed into Block(2,0) as well. This could potentially be explained by the visual distinction between shorter and longer arms in Shirt images, an inter-class variance effect.

\FloatBarrier

\FloatBarrier

\subsubsection{CIFAR 10}

\begin{figure}[!t]
\centering
\includegraphics[width=4.75in]{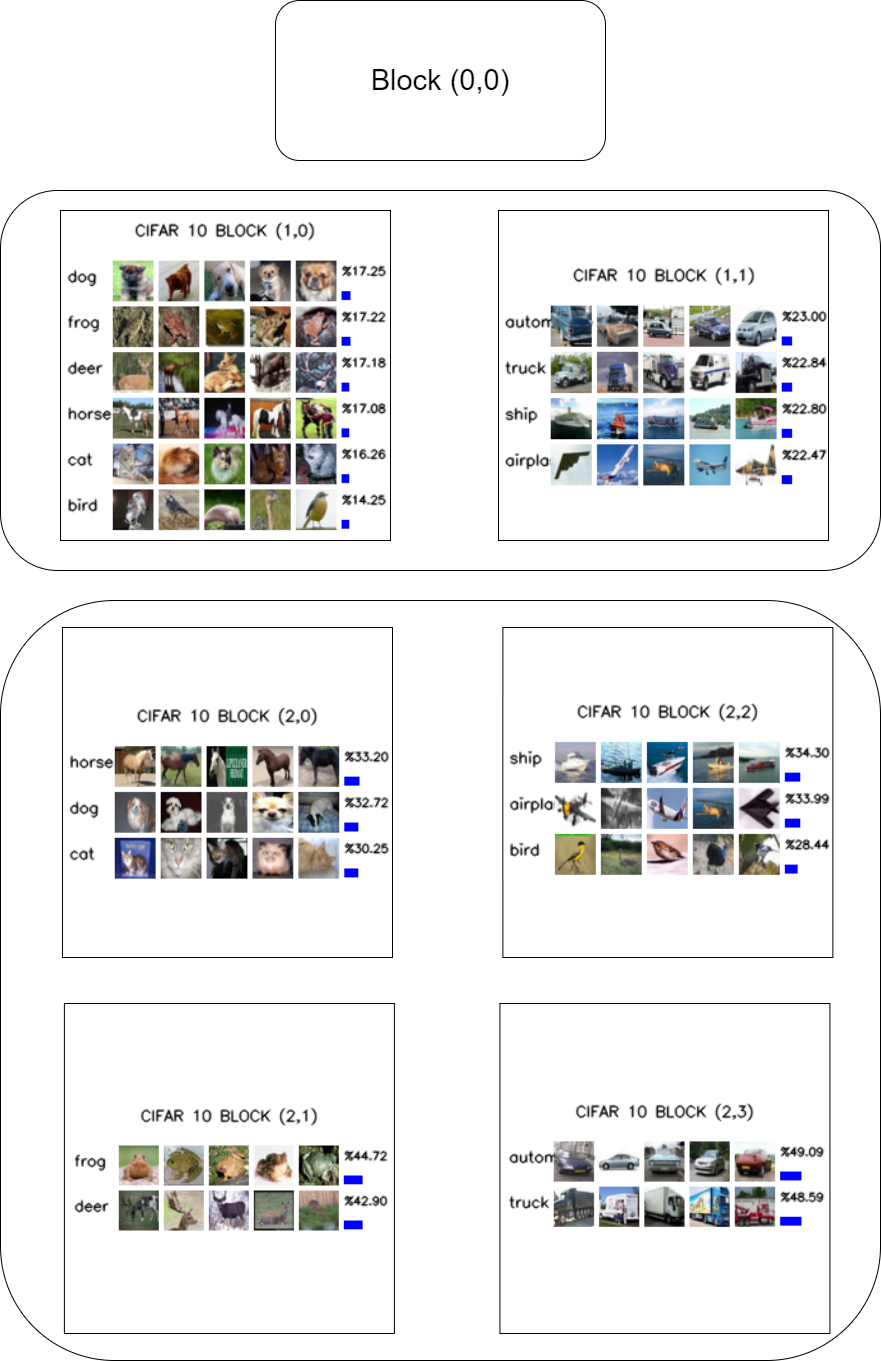}
\caption{The class distribution of CIFAR 10 samples into CIGT routing units.}
\label{CIFAR10_CIGT}
\end{figure}

A grouping of CIFAR 10 test samples induced by the information gain based routing in [1,2,4] CIGT model is shown in Figure \ref{CIFAR10_CIGT}. 

In Block 1, we see that there is a very clear distinction between two groups. Block(1,0) contains animals and artificial, while human made objects (automobile, truck, ship, airplane) are routed into Block(1,1). The visual features of animal classes tend to be similar to each other, compared to human made objects, for example dog, cat and deer classes contain similar four-legged animal shapes. It is also intuitive that road-vehicles like automobile and trucks are sent to into the same computation unit, there are obvious similarities between these classes like the wheels. Ships and airplanes contain sky and the sea as the background that makes those classes visually more similar. It is an interesting observation that more numerous bird samples are routed into  Block(1,1) compared to other animals. A plausible explanation can be that some bird images contain sky as the background and bird images with their wings spread exhibit some degree of visual similarity with the airplane samples. In Block 4, we have four different routes. Block (2,0) contains four legged animals horse, dog and cat dominantly, which is easy to interpret. Block (2,1) contains the classes frog and deer as dominant classes. These classes are not particularly similar to each other, except some background clues with greenish grassy appearance. During CIFAR-10 experiments we have used the balance coefficient as $\lambda_{balance}=5$, so there is a pressure on the routers to maintain a balanced sample distribution among the computation units. This may have lead the router to learn this particular distribution where deer and frog classes are forced into the same route, where it may be semantically more meaningful to send deer samples into Block (2,0) with other four-legged animals and leaving the frog class the single dominant class for Block (2,1). Block (2,2) contains classes ship, airplane and bird as dominant ones. We attribute the routing of bird samples into this unit to the bird-airplane similarity we mentioned before. Finally, Block (2,2) contains automobile and truck classes dominantly, which is easy to interpret semantically.

\begin{figure}
     \centering
     \begin{subfigure}[b]{0.675\textwidth}
         \centering
         \includegraphics[width=\textwidth]{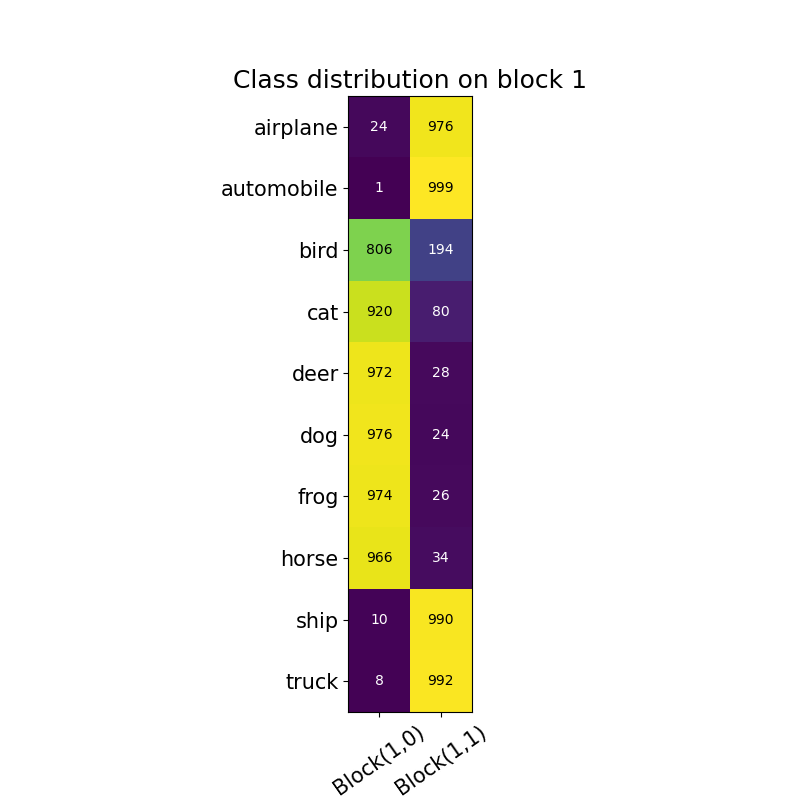}
     \end{subfigure}
     \hfill
     \begin{subfigure}[b]{0.675\textwidth}
         \centering
         \includegraphics[width=\textwidth]{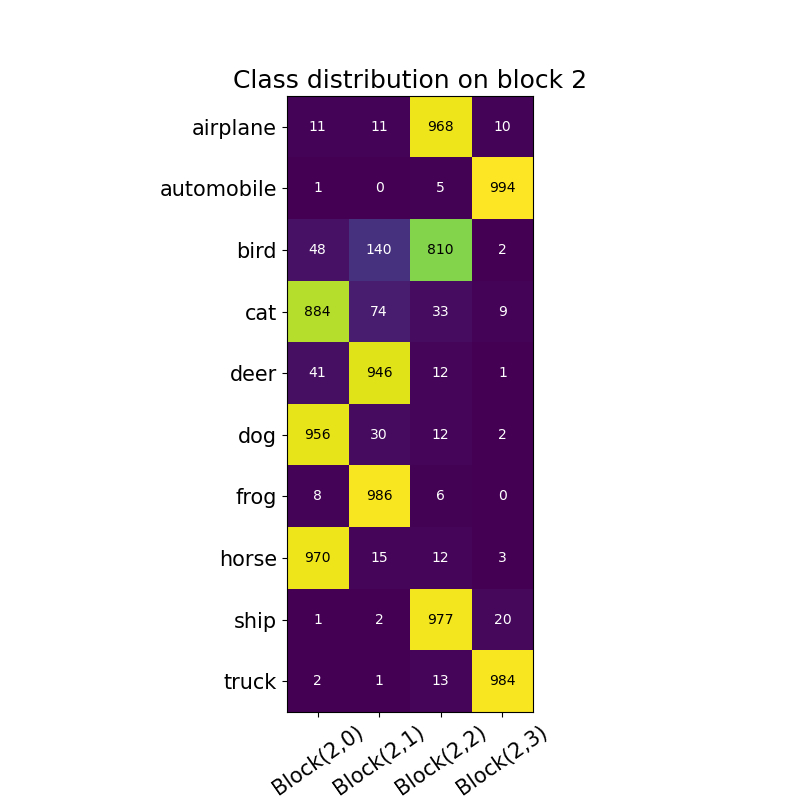}
     \end{subfigure}
        \caption{CIFAR 10 Test set sample distribution in each of routing blocks.}
        \label{fig:CIFAR10_RoutingMatrices}
\end{figure}

In Figure \ref{fig:CIFAR10_RoutingMatrices}, we show the distribution of test samples into different routes with actual numbers in matrix form, which are from the same experiment that has been used for generating Figure \ref{CIFAR10_CIGT}. The numbers support the previous findings and show that semantically similar classes are consistently being routed into the same routes.

\FloatBarrier

\FloatBarrier

\subsubsection{MNIST}
\begin{figure}[!t]
\centering
\includegraphics[width=4.75in]{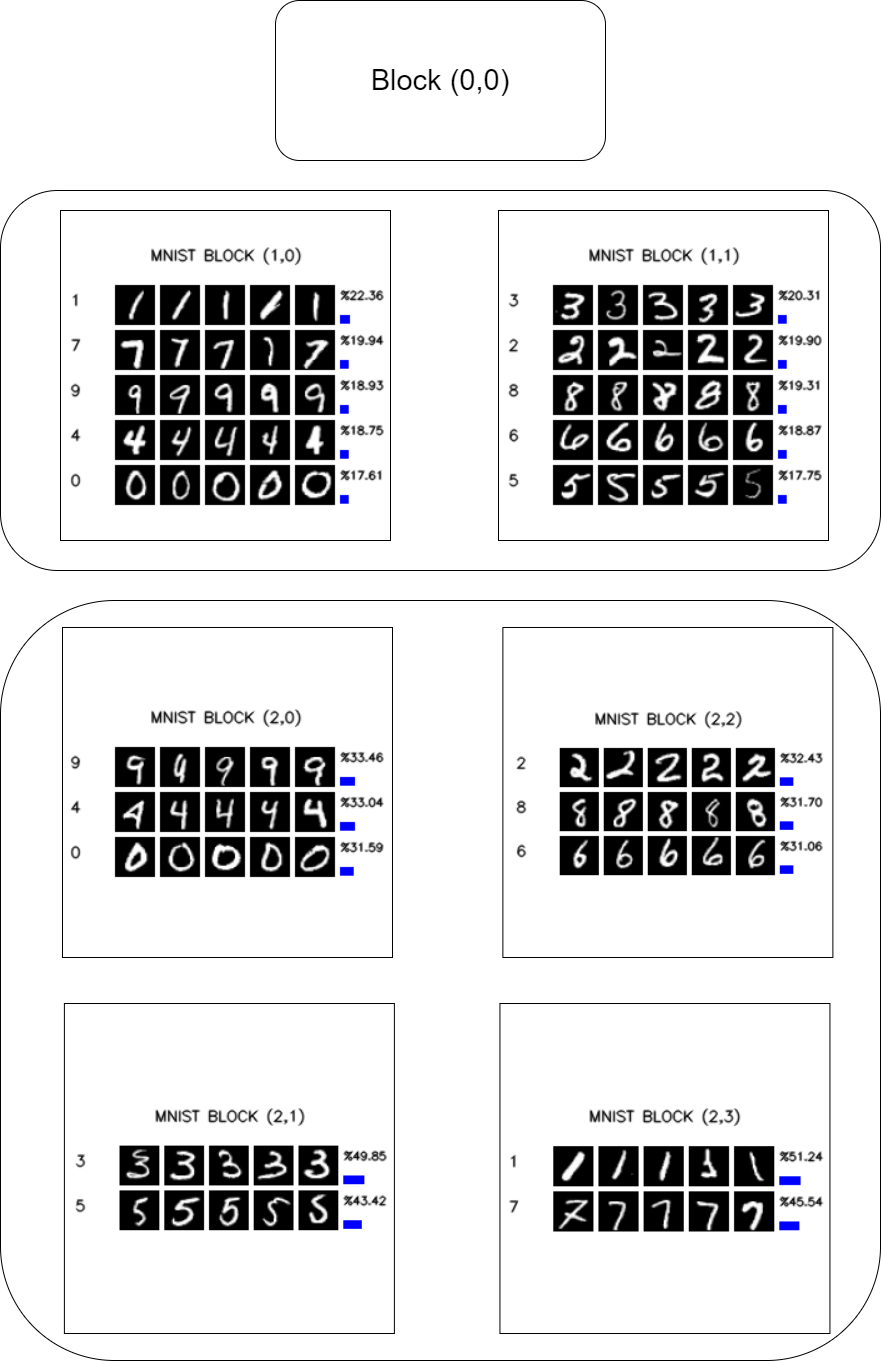}
\caption{The class distribution of MNIST samples into CIGT routing units.}
\label{MNIST_CIGT}
\end{figure}

The CIGT routing of the MNIST dataset on a [1,2,4] architecture is given in Figure \ref{MNIST_CIGT}. In hand-written digits, there is not a clear distinction between class groups from visual grouping point of view. However, we can observe that the CIGT routing divides the classes into groups according to some low level features of the handwritten shapes. For example, the classes 1 and 7 are sent into the same route in both blocks, which is likely due to the linear structure of both numbers. 3 and 5 are routed together, where we have a similar curvature structure, especially in the lower half of the images. The same is true for 4 and 9 as well; the lower part of both classes is drawn similarly.

\begin{figure}
     \centering
     \begin{subfigure}[b]{0.675\textwidth}
         \centering
         \includegraphics[width=\textwidth]{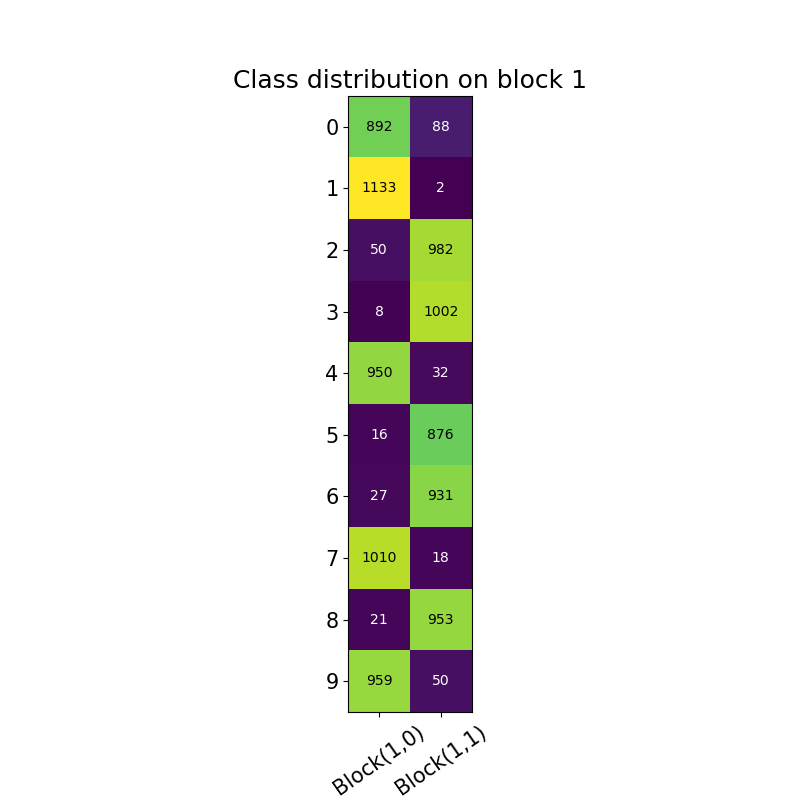}
     \end{subfigure}
     \hfill
     \begin{subfigure}[b]{0.675\textwidth}
         \centering
         \includegraphics[width=\textwidth]{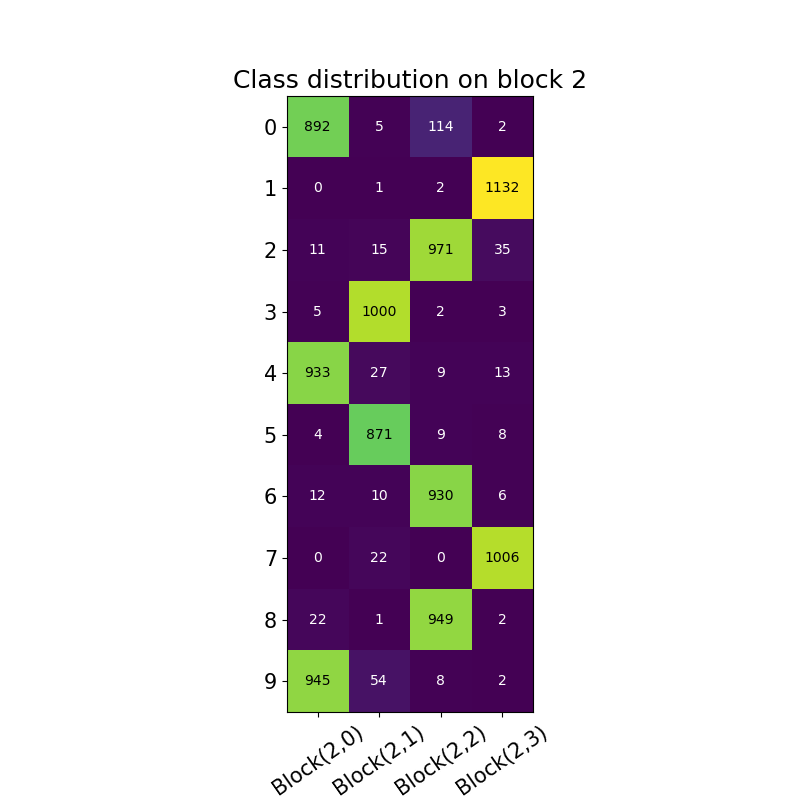}
     \end{subfigure}
        \caption{MNIST Test set sample distribution in each of routing blocks.}
        \label{fig:MNIST_RoutingMatrices}
\end{figure}

Figure \ref{fig:MNIST_RoutingMatrices} gives the numerical distribution of the test samples in both routing blocks. We can see samples from class 0 are sent into Block(2,2) in  significant numbers, in addition to block Block(2,0). In Block(2,2), we have the classes 8 and 6 while in  Block(2,0) we have 9. All these classes have circular features similar to 0 itself. This can explain the tendency of the 0 class samples to get routed into these blocks.

\FloatBarrier

\FloatBarrier

\bibliographystyle{unsrtnat}
\bibliography{cigt_prl}  






\end{document}